\documentclass[conference]{IEEEtran}
\IEEEoverridecommandlockouts
\usepackage{amsmath,amssymb,amsfonts}
\usepackage{algorithmic}
\usepackage{graphicx}
\usepackage{wrapfig}
\usepackage{subfig}
\usepackage{textcomp}
\usepackage{xcolor}

\DeclareMathOperator*{\argmax}{argmax}

\def\BibTeX{{\rm B\kern-.05em{\sc i\kern-.025em b}\kern-.08em
    T\kern-.1667em\lower.7ex\hbox{E}\kern-.125emX}}

\begin{document}

\title{Exploring the Possibility of TypiClust \\for Low-Budget Federated Active Learning
\thanks{© 2025 IEEE.  Personal use of this material is permitted.  Permission from IEEE must be obtained for all other uses, in any current or future media, including reprinting/republishing this material for advertising or promotional purposes, creating new collective works, for resale or redistribution to servers or lists, or reuse of any copyrighted component of this work in other works.}
}

\author{\IEEEauthorblockN{Yuta Ono}
\IEEEauthorblockA{
\textit{The University of Tokyo}\\
Tokyo, Japan \\
ono-yuta116@g.ecc.u-tokyo.ac.jp}
\and
\IEEEauthorblockN{Hiroshi Nakamura}
\IEEEauthorblockA{
\textit{The University of Tokyo}\\
Tokyo, Japan \\
nakamura@hal.ipc.i.u-tokyo.ac.jp}
\and
\IEEEauthorblockN{Hideki Takase}
\IEEEauthorblockA{
\textit{The University of Tokyo}\\
Tokyo, Japan \\
takasehideki@hal.ipc.i.u-tokyo.ac.jp}
}
\maketitle

\begin{abstract}
Federated Active Learning (FAL) seeks to reduce the burden of annotation under the realistic constraints of federated learning by leveraging Active Learning (AL).
As FAL settings make it more expensive to obtain ground truth labels, FAL strategies that work well in low-budget regimes, where the amount of annotation is very limited, are needed.
In this work, we investigate the effectiveness of TypiClust, a successful low-budget AL strategy, in low-budget FAL settings.
Our empirical results show that TypiClust works well even in low-budget FAL settings contrasted with relatively low performances of other methods, although these settings present additional challenges, such as data heterogeneity, compared to AL.
In addition, we show that FAL settings cause distribution shifts in terms of typicality, but TypiClust is not very vulnerable to the shifts.
We also analyze the sensitivity of TypiClust to feature extraction methods, and it suggests a way to perform FAL even in limited data situations.\footnote{Accepted at COMPSAC 2025}
\end{abstract}


\section{Introduction}
Recent years have witnessed significant progress in deep learning for image classification~\cite{Krizhevsky2012-uy, He2015-wd, Dosovitskiy2020-dv, Alayrac2022-xv}.
Training a deep learning model to achieve high classification accuracy often necessitates abundant annotated images gathered in one place to perform fully-supervised training in a centralized way.
However, this requirement frequently poses a challenge when we utilize these classification methods in real-world settings.
In practice, annotating data is both time-consuming and costly, and datasets obtained by distributed clients cannot be shared due to privacy concerns.
One of the ways to tackle this problem is federated active learning~(FAL).
Federated active learning is a combination of active learning~(AL) and federated learning~(FL).
It aims to achieve high performance with fewer labeled data points by clients selecting the most informative unlabeled data points to be annotated for FL.

What differentiates FAL from AL the most is a restriction on data-sharing among clients.
This restriction sometimes forces us to perform FAL with heterogeneous datasets, making conventional AL methods useless.
To this end, some FAL-specific strategies have been proposed~\cite{Cao2022-ej, Kim2023-fw, Chen2023-dk}.
The restriction poses another challenge to FAL: more expensive labels.
Nowadays, it is common to crowdsource the annotation task to obtain ground truth inexpensively and quickly.
However, crowdsourcing the annotation is impossible in FAL due to the data-sharing restriction, leading to more expensive labels.

One solution to the expensive label problem is low-budget FAL, which seeks to maximize model performance with minimum annotation cost.
Low-budget FAL tries to optimize a model through FAL, keeping the amount of annotation small.
As FAL is an emerging technology and the prior works focus on relatively high-budget regimes, the effectiveness of FAL in low-budget regimes remains unexplored.
The performance of existing FAL methods in low-budget regimes needs to be evaluated because FAL methods that work well in low-budget settings will facilitate the use of deep learning in more realistic scenarios.

In this work, we focus on TypiClust, a successful low-budget AL strategy.
Although TypiClust has proven effective in low-budget AL settings, it remains unclear whether it works well in FAL settings because FAL involves heterogeneous or small datasets, making self-supervised learning challenging, on which TypiClust highly depends.
For this reason, we explore the possibility of TypiClust in low-budget FAL settings.

Our experimental evaluation highlights TypiClust's remarkable performance in low-budget FAL settings compared to baseline methods, including ones tailored for FAL settings.
In addition, our analysis of TypiClust's sensitivity to feature spaces reveals that TypiClust is not vulnerable to feature extraction methods.
This suggests using pre-trained models for feature extraction, which allows clients with an extremely limited amount of unlabeled data to participate in FAL.

\section{Preliminaries}
\textbf{Federated Learning (FL)}
is a distributed machine learning paradigm that aims to train a global model iteratively without sharing raw data owned by clients. 
In FL, each client performs model training using data possessed by the client and sends information about model updates, e.g., gradients, to the central server.
The central server aggregates the information gathered from clients to update a global model.
The updated global model is distributed to clients, and then each client resumes training.
Since FL does not require clients to share raw data, we can train a model even with privacy-sensitive data, such as medical images, that clients cannot share.

The minimization objective in FL can be formulated as follows:
\begin{equation}
    f(w) = \sum_{k=1}^{K}\frac{|\mathcal{P}_k|}{\sum_k |\mathcal{P}_k|} F_k(w)
\end{equation}
where 
\begin{equation}
    F_k(w) = \frac{1}{|\mathcal{P}_k|}\sum_{i \in \mathcal{P}_k}f_i(w),
\end{equation}
\(\mathcal{P}_k\) is a data partition that belongs to the client \(k\) and \(f_i(w) = l(x_i, y_i; w)\) is the loss of an example \((x_i, y_i)\) with model parameters \(w\).
In this work, the term ``non-IID datasets'' or ``heterogeneous datasets'' shall be used to mean \(\mathbb{E}_{\mathcal{P}_k}\left[ F_k(w) \right] \neq f(w)\)~\cite{McMahan2017-pc}.

In FL, each client trains a model using a local dataset to minimize the local loss.
\begin{equation}
    w_k^{(r+1)} = w_{\mathrm{global}}^{(r)} - \eta \nabla F_k(w^{(r+1)}_k),
\end{equation}
where \(w_k^r\) is the local model parameters in the client \(k\) at round \(r\).
The updated local model parameters are shared with the central server, and information from all the clients is aggregated to obtain the global model for the next round \(w^{(r+1)}_{\mathrm{global}}\).
\begin{equation}
    w_{\mathrm{global}}^{(r+1)} = \mathrm{Aggregate}(\{w_k^{(r)}\}_{k=1}^{K})
\end{equation}

\textbf{Active Learning (AL)}
mitigates the annotation cost by querying the most informative samples from an unlabeled data pool.
Active learning strategies can be categorized into uncertainty-based and diversity-based.
Uncertainty-based strategies utilize the prediction uncertainty of the classification model under training to select data valuable for model improvement.
They tend to underperform when acquiring a batch of data points in one go, selecting similar data points simultaneously because similar data points have similar uncertainties.
Diversity-based ones aim to annotate diverse sets of instances to cover the input space efficiently, avoiding the selection of too similar samples, which is often redundant and useless for model training.
Hybrid strategies also exist, which try to query uncertain instances covering a wide range of the input space.

Here, we describe the procedure of AL.
A client have an unlabeled dataset \(U_r = \{x_i\}_{i=1}^{|U_r|}\) and a labeled dataset \(L_r = \{(x_i, y_i)\}_{i=1}^{|L_r|}\) at round \(r\).
We train a model using the labeled dataset and update model parameters to \(w_r\).
After the training, the client selects a data point to be annotated from the unlabeled dataset.
To select the data, we use an acquisition function \(A(x; w_r)\), which is designed to estimate the informativeness of an unlabeled data point:
\begin{equation}
  x^\star_r = \argmax_{x \in U_r} A(x; w)
\end{equation}
where \(x^\star_r\) is the data point selected for annotation at round \(r\).
Note that we usually select a set of data every round because it can reduce the number of times the model is retrained~\cite{Kirsch2019-pv}.
The selected unlabeled data point is annotated and then added to the labeled dataset:
\begin{equation}
    L_{r+1} = L_r \cup \{(x^\star_r, y^\star_r)\},
\end{equation}
\begin{equation}
    U_{r+1} = U_r \setminus \{x^\star_r\}.
\end{equation}
Following this procedure, the model is trained iteratively while the budget for annotation remains.

\textbf{Federated Active Learning (FAL)} has been studied to leverage FL in more realistic scenarios.
The restriction on data sharing in FL often requires us to perform training with heterogeneous datasets.
It is strongly needed to mitigate the effect of heterogeneity to achieve better performance.
To this end, various FAL strategies have been proposed~\cite{Cao2022-ej, Kim2023-fw, Chen2023-dk}.
These strategies leverage properties specific to FAL settings (e.g., model disagreement between global and local models) to deal with the data heterogeneities in FAL.

In FAL, we first perform an AL round in each client, i.e., each client selects a set of unlabeled data from the unlabeled dataset of the client and then annotates them.
After annotation, each client independently trains a model from the aggregated model at the previous FL round.
The trained model is sent to the central server and aggregated to obtain a global model.

\section{Methodology}
To begin with, we clarify the meaning of the term \textit{low budget}.
It indicates that the initial labeled set for model training is small or empty and that we have a limited budget for annotation.
In other words, we are only able to annotate very limited data points (\(\approx\) number of classes) in an iteration of AL/FAL to train a classification model from scratch.

TypiClust is an AL strategy suited for low-budget regimes~\cite{Hacohen2022-lr}.
It is theoretically justified and empirically observed that TypiClust performs well in low-budget AL regimes.
Initially, TypiClust performs self-supervised learning, including SimCLR~\cite{Chen2020-dj, Chen2020-uf} and DINO~\cite{Caron2021-al, Oquab2023-cz}, as a pretext task to obtain informative representation from unlabeled data.
In this work, we use SimCLR.
After obtaining features by self-supervised learning, ``typicalities'' of data points are calculated in the feature space.
Typicality is defined as follows:
\begin{equation}
    Typicality(z) = \left( \frac{1}{K}\sum_{z_i \in K\mathrm{-NN}(z)}\|z - z_i\|_2 \right)^{-1}.
\end{equation}
Here, \(z\ (\in \mathbb{R}^d)\) is a data point in the extracted feature space and \(K\mathrm{-NN}(z)\) denotes a set of \(K\) nearest neighbors of \(z\) in the feature space.
TypiClust builds clusters in the feature space at every iteration of AL and selects the most typical instance from each cluster.
As TypiClust can select informative data points to be annotated independently with the classifier under training, it is not affected by the unstable performance of the classifier at the beginning of AL and in low-budget AL.

In this work, we investigate the effectiveness of TypiClust, focusing on low-budget FAL, mainly targeting practical FAL in cross-silo FL~\cite{Huang2022-nj}, albeit not limited to it.
In the experiments, we assume distributed machine learning where data cannot be shared with other clients due to privacy concerns or communication costs, and the budget for annotation is very limited.
We also study the possibility of applying FAL to the cross-device case (e.g., distributed machine learning with remote sensors), by considering the potential of exploiting pre-trained models for FAL with limited computational resources.

\section{Results}
\subsection{Experimental setup}
\subsubsection{Datasets}
\begin{figure}[tb]
  \begin{minipage}[t]{0.48\columnwidth}
    \centering
    \includegraphics[width=\columnwidth]{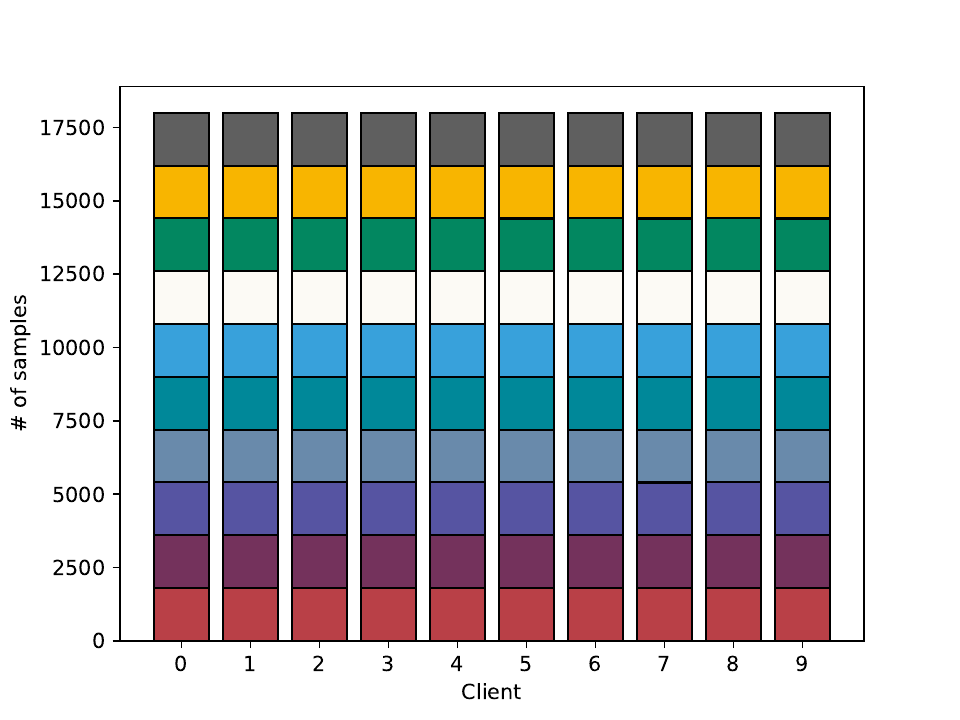}
    \caption{CINIC-10 (\(\alpha=\infty\))}
    \label{fig:cinic-inf}
  \end{minipage}
  \hspace{0.04\columnwidth}
  \begin{minipage}[t]{0.48\columnwidth}
    \centering
    \includegraphics[width=\columnwidth]{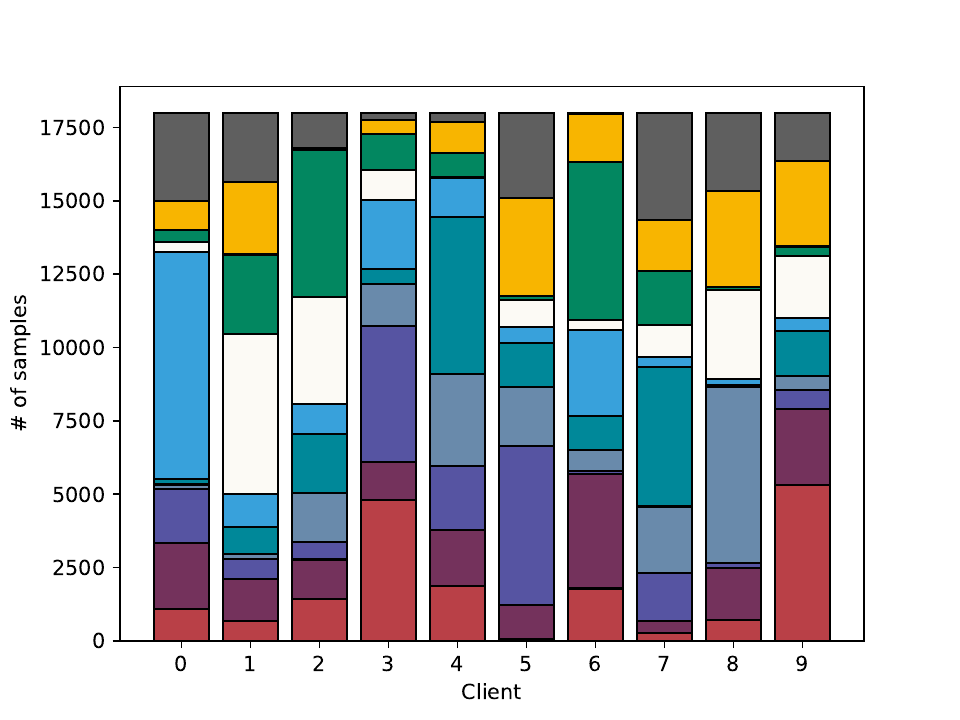}
    \caption{CINIC-10 (\(\alpha=1.0\))}
    \label{fig:cinic-1}
  \end{minipage} \\

  \begin{minipage}[t]{0.48\columnwidth}
    \centering
    \includegraphics[width=\columnwidth]{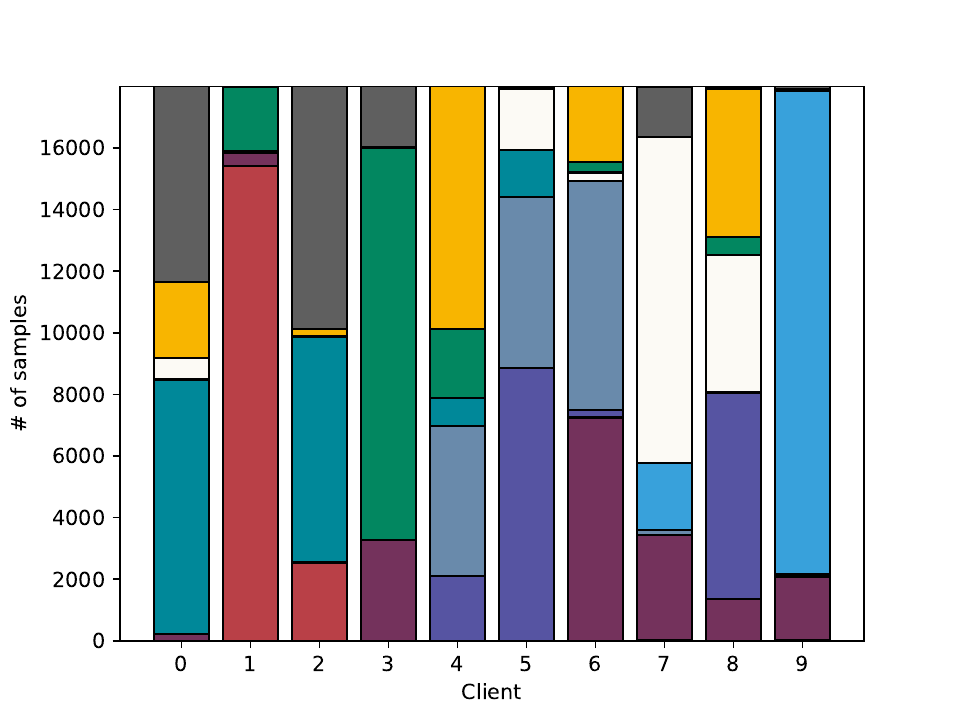}
    \caption{CINIC-10 (\(\alpha=0.1\))}
    \label{fig:cinic-01}
  \end{minipage}
  \hspace{0.04\columnwidth}
  \begin{minipage}[t]{0.48\columnwidth}
    \centering
    \includegraphics[width=\columnwidth]{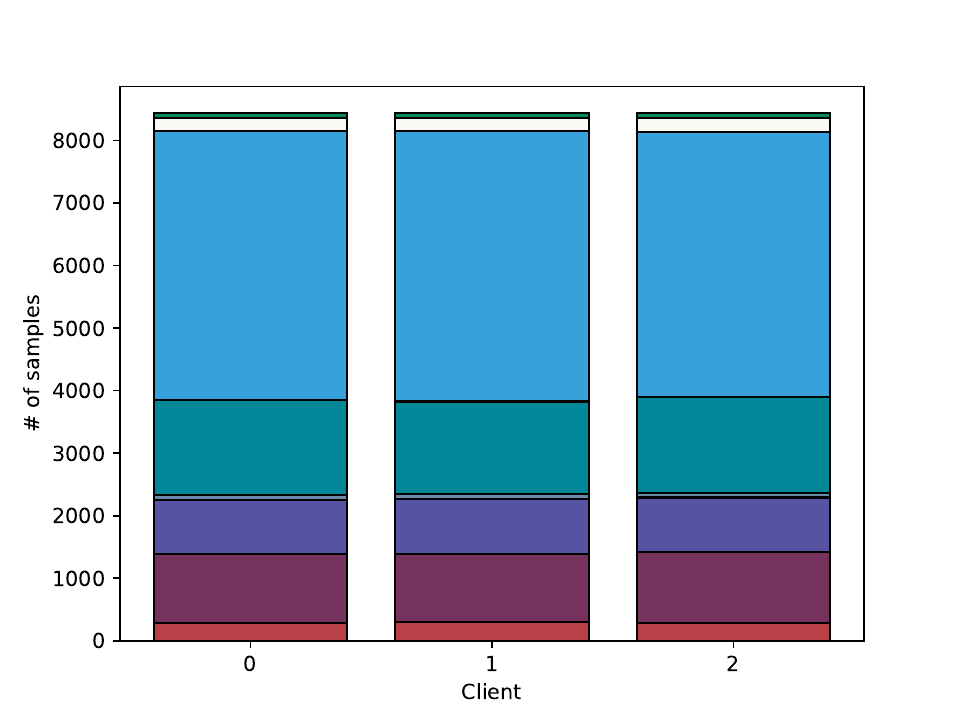}
    \caption{ISIC2019}
    \label{fig:isic-random}
  \end{minipage}
\end{figure}
All methods are evaluated on image classification tasks using CINIC-10~\cite{Darlow2018-se} and ISIC2019~\cite{Kassem2020-cg}.
CINIC-10 consists of images from CIFAR-10~\cite{Krizhevsky2009-jz} and ImageNet~\cite{Deng2009-fk} with 10 classes.
ISIC2019 is an imbalanced skin-lesion image dataset with eight classes.
To simulate non-IID data distributions, we utilize the Latent Dirichlet Allocation strategy~\cite{Kim2023-fw,Li2021-rp}.
We allocate data to client \(k\) by sampling from \(p_k \sim \mathrm{Dir}(\alpha \cdot \vec{1})\), where \(\vec{1} \in \mathbb{R}^C\).
CINIC-10 is partitioned for ten clients with three different \(\alpha\) values \(\alpha=0.1, 1.0, \infty\) (See Figure~\ref{fig:cinic-inf}, Figure~\ref{fig:cinic-1}, and Figure~\ref{fig:cinic-01}; each color is responding to a class).
ISIC2019 is partitioned uniformly at random for three clients (See Figure~\ref{fig:isic-random}).
ISIC2019 has an imbalance ratio larger than 50, meaning a 50-fold difference in the number of images between the most and least frequent classes.
Different parts of the partitioned dataset are then distributed to different clients.
The number of clients in FAL is set to 10 for CINIC-10 and 3 for ISIC2019, considering real-world applications and dataset sizes.

\subsubsection{Budgets}
We use tiny and small budget sizes for active querying that involve query step sizes of 1 and 3 times the number of classes per client, respectively.

\subsubsection{Classifiers}
As a classifier that is trained by FAL procedures, we choose two models, a simple CNN (cnn-4) and ResNet18~\cite{He2015-wd}.
The simple CNN comprises four convolutional layers followed by a fully connected layer.

\subsubsection{Baselines}
We consider five conventional AL strategies and two FAL strategies.
\textbf{Random} selects instances to be annotated uniformly at random.
\textbf{Entropy} and \textbf{margin} select uncertain instances using the trained classifier's outputs.
\textbf{Coreset}~\cite{Sener2017-hm} and \textbf{BADGE}~\cite{Ash2020-wr} try to acquire diverse samples that represent a feature space well.
Coreset uses the embedding space generated by the penultimate layer of the classifier, and BADGE works on the gradient embedding space.
All the conventional methods, excluding random sampling, perform sampling with two options for a query selector in FAL settings.
They can choose the global model or local-only model, which is trained only with a local labeled dataset without communicating with other clients, for calculating uncertainty and diversity metrics.
\textbf{KAFAL}~\cite{Cao2022-ej} is a strategy tailored for FAL.
It prioritizes sampling data points on which the global and local-only models disagree.
\textbf{LoGo}~\cite{Kim2023-fw} consists of two steps, macro and micro steps.
In the macro step, clusters are built in a gradient embedding space by the local-only model to ensure the diversity of instance selection, and in the micro step, the most uncertain data point is selected from each cluster.

\subsubsection{Configurations}
\begin{table}[b]
\caption{Experimental configurations}
\label{tab:configs}
\centering
\begin{tabular}{cccc}
\hline
Dataset & \(\mathrm{Dir}(\alpha)\) & Model & Budget \\\hline
CINIC-10 & 0.1 & Simple CNN & tiny \\
CINIC-10 & 1.0 & Simple CNN & tiny \\
CINIC-10 & \(\infty\) & Simple CNN & tiny \\
CINIC-10 & \(\infty\) & Simple CNN & small \\
CINIC-10 & 0.1 & ResNet18 & tiny \\
CINIC-10 & 1.0 & ResNet18 & tiny \\
CINIC-10 & \(\infty\) & ResNet18 & tiny \\
CINIC-10 & \(\infty\) & ResNet18 & small \\
ISIC2019 & -- & ResNet18 & tiny \\
ISIC2019 & -- & ResNet18 & small \\
\hline
\end{tabular}
\end{table}
\begin{table}[b]
\caption{Hyperparameters for experiments}
\label{tab:hyper}
\centering
\begin{tabular}{ll}
\hline
Learning rate & 0.01    \\
Optimizer     & SGD     \\
Momentum      & 0.9     \\
Weight decay  & 0.0001  \\
\#clients & 10 (CINIC-10) / 3 (ISIC2019) \\
FL algorithm & FedAvg \\
Local epochs  & 10      \\
Global rounds & 10      \\\hline
\end{tabular}
\end{table}

We evaluate each FAL strategy in ten configurations, changing datasets, heterogeneity levels, classifier models, and budget sizes as shown in Table~\ref{tab:configs} with four different random seeds.
In each FAL round, local models are trained for 10 epochs using the local datasets, and the model information is aggregated by FedAvg~\cite{McMahan2017-pc}.
Other hyperparameters for training are shown in Table~\ref{tab:hyper}.

\subsection{Main results}
\label{sec:main-results}
\begin{figure*}[tb]
    \centering
    \includegraphics[width=\linewidth]{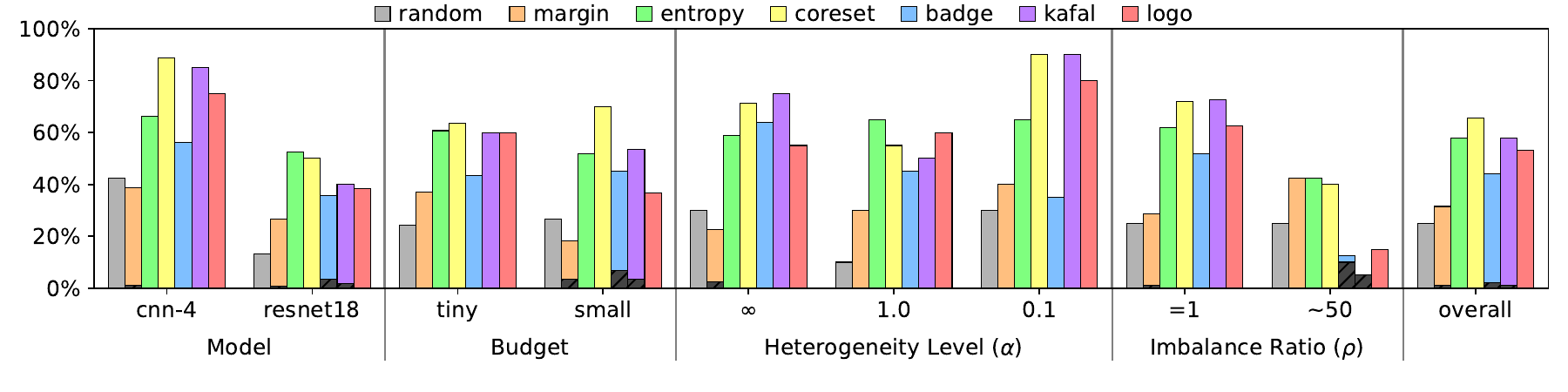}
    \caption{Win rates (colored bars) and defeat rates (black hatched bars) of Typiclust against baselines. Higher values of win rate and lower values of defeat rate imply the superiority of TypiClust.}
    \label{fig:results}
\end{figure*}

We evaluate FAL strategies by using the \(t\)-test framework, a widely accepted framework for conventional AL and recent FAL literature~\cite{Kim2023-fw,Ash2020-wr,Parvaneh2022-zn}.
In this framework, we first run each FAL strategy with four different random seeds and obtain \(a_{r,l}^{i}\), which is a performance metric value (e.g., test accuracy) of the strategy \(i\) at round \(r\) with \(l\)-th random seed.
The \(t\)-score of two strategies \(i\) and \(j\) at the round \(r\) (\(=t^{ij}_r\)) is calculated using the mean and standard variance of the difference between two classification performances over all the trials with four different random seeds as follows:
\begin{equation}
    t^{ij}_r = \frac{\sqrt{4}{\mu^{ij}_r}}{\sigma^{ij}_{r}}
\end{equation}
where
\begin{equation}
 \mu^{ij}_r = \frac{1}{4}\sum_{l=1}^{4} \left( a_{r,l}^{i} - a_{r,l}^{j}\right)
\end{equation}
and
\begin{equation}
    \sigma_r^{ij} = \sqrt{\frac{1}{3} \sum_{l=1}^{4} \left[\left( a_{r,l}^{i} - a_{r,l}^{j} \right) - \mu_r^{ij}\right]^2}.
\end{equation}
The strategy \(i\) is regarded to defeat the strategy \(j\) if \(t^{ij}_r > 2.776\).
We evaluate the superiority of TypiClust over every baseline by conducting a pair-wise two-sided \(t\)-test.
Note that we use balanced recall on ISIC2019 for performance evaluation because the test dataset of ISIC2019 is imbalanced.
The win rate of the strategy \(i\) over the strategy \(j\) is calculated as follows:
\begin{equation}
    p_{ij} = \frac{1}{R} \sum_{r=1}^{R} 1_{t_r^{ij} > 2.776}.
\end{equation}

\begin{figure}[tb]
    \centering
    \includegraphics[width=\columnwidth]{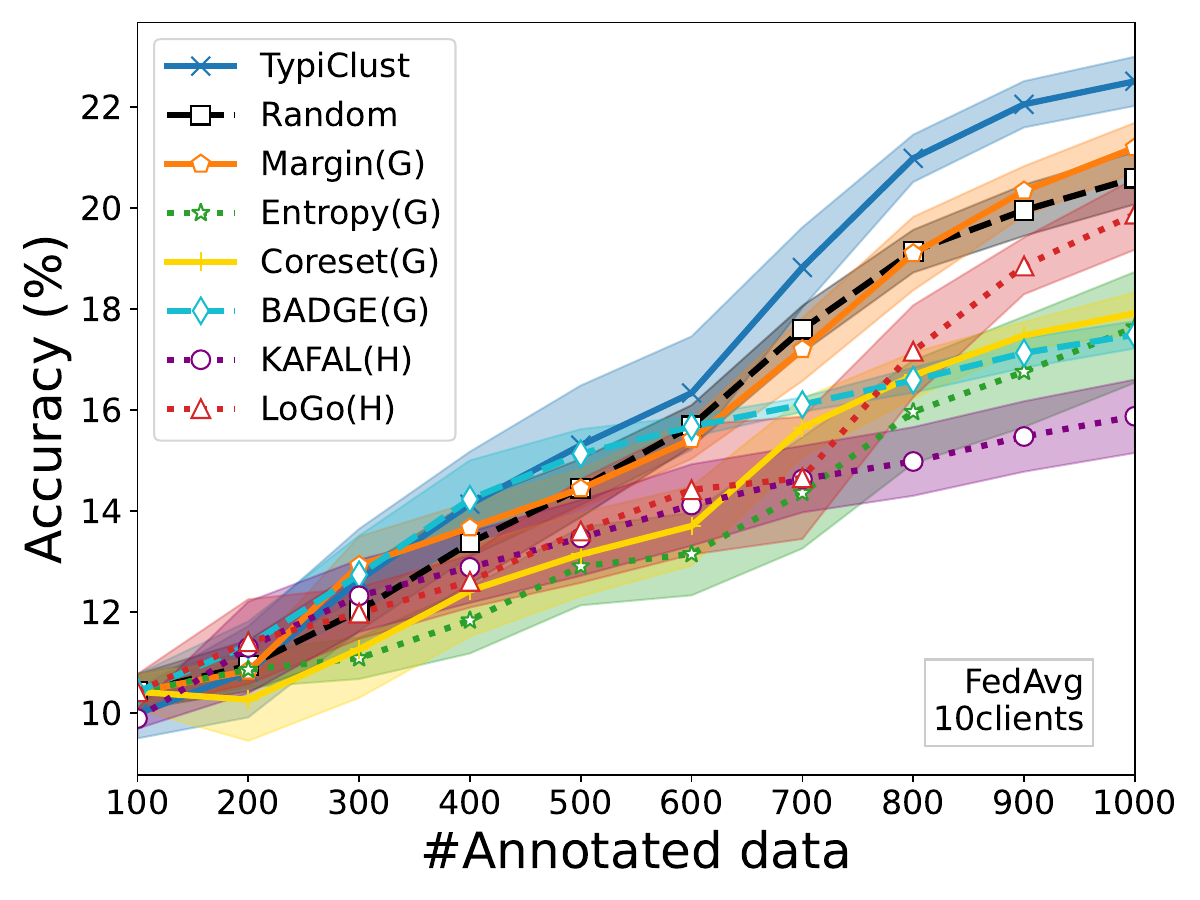}
    \caption{Accuracies of ResNet18 on CINIC-10 with heterogeneity \(\alpha=\infty\) in the tiny budget regime. The mean and standard error with four different random seeds are shown. We observe that no method other than TypiClust outperforms random sampling by a large margin, concluding that the cold-start problem also occurs in FAL settings.}
    \label{fig:cold-start}
\end{figure}

We present the main results of our evaluation in Figure \ref{fig:results}.
Longer color bars imply the higher win rates of TypiClust over baselines.
Overall, TypiClust shows significant performance even in FAL settings, being underscored by win rates that are higher than defeat rates across all the baselines.
Notably, TypiClust shows the lowest win rate against random sampling.
This suggests the existence of the cold-start problem in FAL (See also Figure~\ref{fig:cold-start} for an example), meaning other baselines can be defeated by random sampling in low-budget FAL.

Focusing on the results grouped by the classification models, TypiClust realizes higher win rates with the simple CNN than the ResNet-18.
This is considered to be because the simple model can be trained well even with a small dataset and contrasts the performance differences between strategies.

TypiClust shows great performance in both tiny-budget and small-budget regimes and three different heterogeneity levels.
When the local inter-class balance collapses, i.e., \(\alpha=0.1\), TypiClust's win rates reach the highest values against most baselines.

To summarize, TypiClust outperforms baselines in low-budget settings, and its advantage over baselines is more significant with a simple model and heterogeneous data partitions.
In addition, interestingly, the second-best method is the random selection strategy, which suggests the existence of the cold-start problem~\cite{Hacohen2022-lr,Nguyen2015-ru,Gal2015-sy} in FAL as well as AL.

\subsection{Typicality distribution shift}
\begin{figure}[tb]
    \centering
    \includegraphics[width=\columnwidth]{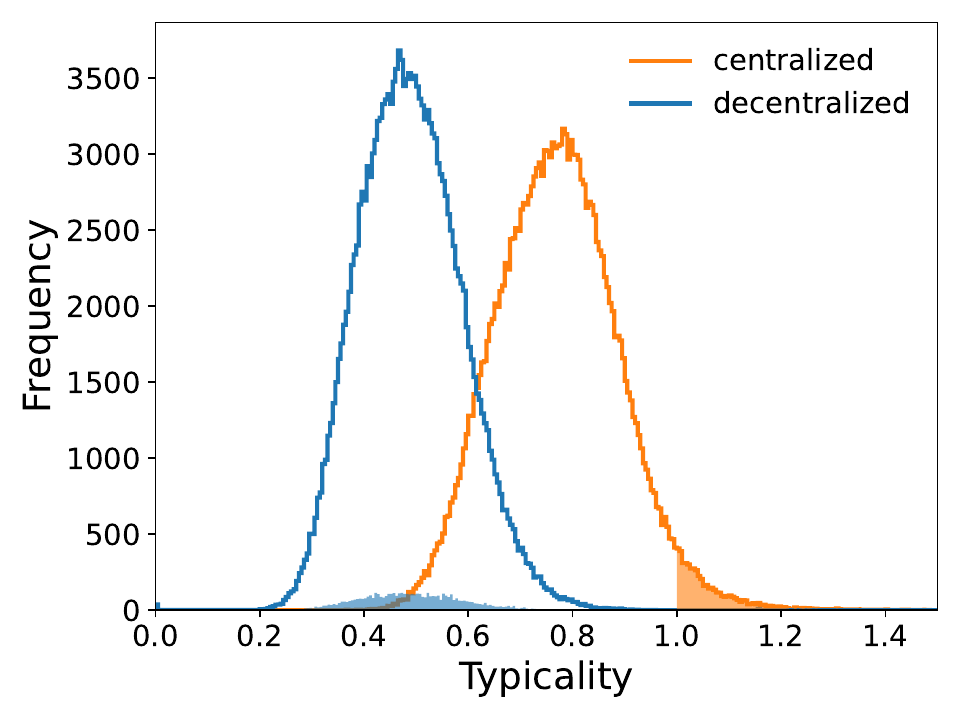}
    \caption{Distribution shift of Typicality in CINIC-10. The filled area comes from the samples with typicalities over 1.0 in a centralized dataset. Typical features in a centralized self-supervised feature space are not always typical in decentralized counterparts in FAL.}
    \label{fig:typicality-distribution}
\end{figure}

It is predicted that the distributions of typicality in AL and FAL settings are different, even if we use the same dataset and self-supervised learning method.
In FAL, we need to perform self-supervised learning and calculate typicality separately in each client, as the data held by a client cannot be shared with other clients.
This separated self-supervised learning can lead to different distributions of typicality and unaligned embeddings.

Although we observed that TypiClust works well with self-supervised features extracted in a decentralized way, it remains unclear if the data distribution in the embedding space varies among clients.
It is important to observe how the distribution of typicality shifts and how the shift affects the performance of TypiClust.
As typicality-based methods~\cite{Hacohen2022-lr,Ono2024-ad,Okano2025-tw} heavily depend on the typicality computed in the embedding space, a distribution shift of typicality can lower their performance.
Thus, we test whether the setting of distributed self-supervised training in FAL shifts the distribution of typicality.

Figure~\ref{fig:typicality-distribution} shows the distribution of typicalities in a centralized dataset of AL and decentralized counterparts of FAL.
Although those datasets consist of the same instances, typicality has different distributions.
Decentralized feature extraction shifts the distribution to the left, meaning that separating one dataset to clients makes the distribution in the embedding space sparse and leads to lower typicalities.
Moreover, the typical samples in the centralized dataset are not necessarily typical in decentralized datasets.
These are considered to be because of the data heterogeneity.
Interestingly, despite this distribution shift, TypiClust works well in FAL settings.

\subsection{Sensitivity analysis to feature spaces}
\label{sec:feature-space}
\begin{figure}
    \centering
    \includegraphics[width=\columnwidth]{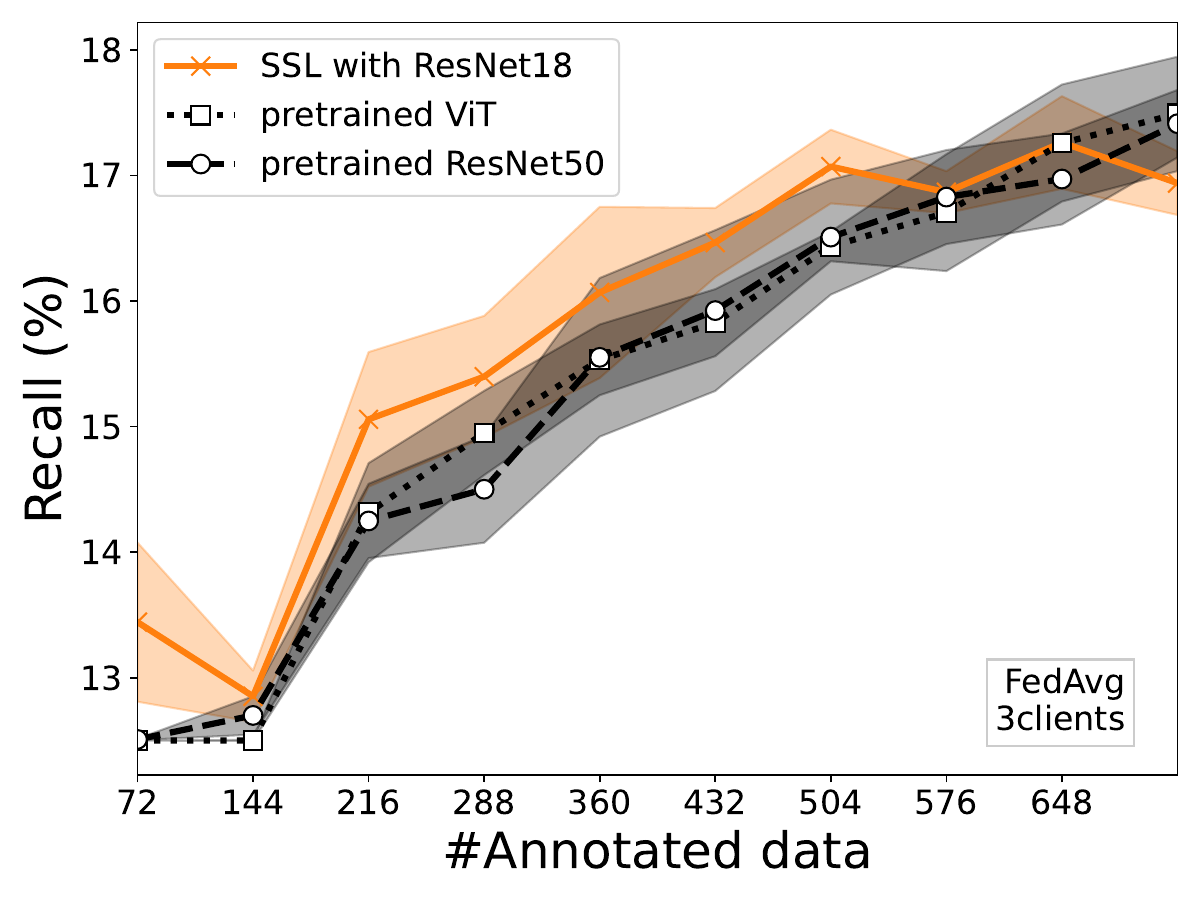}
    \caption{Comparison of TypiClust performances with different feature spaces on ISIC2019 in the small budget regime.}
    \label{fig:ssl-small}
\end{figure}
We cannot straightforwardly apply TypiClust to scenarios where not only the annotation budget but also the amount of unlabeled data is limited.
TypiClust is highly dependent on self-supervised features, but self-supervised learning does not work well and fails to obtain good embeddings with a small amount of data.
In such scenarios, it is helpful if features that are extracted by a pre-trained model can substitute for the self-supervised features.
There are various pre-trained models publicly available, and these models are expected to be able to extract compressed features from data without additional training.

We compare TypiClust's performance using three different feature spaces: SSL features, features extracted by a pre-trained ViT and a pre-trained ResNet50.
These two pre-trained models are trained on ImageNet and are publicly available in PyTorch~\cite{paszke2017automatic}.
Figure~\ref{fig:ssl-small} shows that TypiClust with features extracted by pre-trained models performs as well as that with self-supervised features, indicating the potential of using pre-trained models for TypiClust.

\section{Conclusion}
Low-budget FAL is one of the key frameworks to exploit deep learning in more realistic and practical scenarios, where data instances are distributed to clients and cannot be shared with other clients due to privacy concerns.
In this work, we focused on TypiClust, a well-established AL strategy for low-budget regimes, as a promising approach in low-budget FAL settings.
Our empirical experiment revealed that TypiClust performs well even in such settings with different obstacles from conventional AL settings.
Our findings facilitate the broader applications of low-budget FAL in the real world.

Although TypiClust works better than other baselines in low-budget FAL, there should be room for improvement.
To achieve higher performance, exploiting self-supervised features for model training can be a future direction of research.
However, we cannot na\"ively use the features to train models because the features extracted by different clients are not necessarily aligned, meaning even the same data instance can have different embeddings in different clients.
Methods to align self-supervised features from different clients during or after self-supervised learning are needed.

Federated active learning settings contain more hyperparameters than normal AL, which can be controlled depending on scenarios, such as aggregation algorithms for federated learning, budget size, and local training epochs.
Since AL is known to be sensitive to hyperparameters, changing the hyperparameters is also supposed to vary FAL performance significantly.
Although it is essential to reveal the sensitivity for efficiently utilizing FAL in practice, we leave it as future work.

\section*{Acknowledgments}
A part of this work was supported by JST CREST Grant Number JPMJCR21D2, Japan.

\bibliographystyle{IEEEtran}
\bibliography{reference}
\end{document}